\documentclass[10pt,twocolumn,letterpaper]{article}

\usepackage[pagenumbers]{cvpr} 

\definecolor{cvprblue}{rgb}{0.21,0.49,0.74}
\usepackage[pagebackref,breaklinks,colorlinks,allcolors=cvprblue]{hyperref}
\usepackage{multirow}
\usepackage{adjustbox}
\usepackage{bbm}
\usepackage{bm}
\usepackage{amsthm,amsmath,amssymb}
\usepackage{mathrsfs}

\def\paperID{6333} 
\def\confName{ICCV}
\def\confYear{2025}

\newcommand{\best}[1]{\textbf{#1}}
\newcommand{\second}[1]{\underline{#1}}

\newcommand{\name}{\text{VicaSplat}}
\newcommand{\NAME}{\textbf{\name}}

\title{{\name}: A Single Run is All You Need for 3D Gaussian Splatting and \\ Camera Estimation from Unposed Video Frames}

\author{Zhiqi Li$^{1,2}$\qquad Chengrui Dong$^{1,2}$ \qquad Yiming Chen$^{1,2}$ \qquad Zhangchi Huang$^{1,2}$ \qquad Peidong Liu$^{2,\dag}$\vspace{0.1cm} \\
 $^{1}$Zhejiang University 
 \qquad 
 $^{2}$Westlake University \\
{\tt\small \{lizhiqi49, dongchengrui, chenyiming, huangzhangchi, liupeidong\}@westlake.edu.cn}
}

\let\oldtwocolumn\twocolumn
\renewcommand\twocolumn[1][]{%
    \oldtwocolumn[{#1}{
    \begin{center}
           \vspace{-10pt}
           \includegraphics[width=0.95\linewidth]{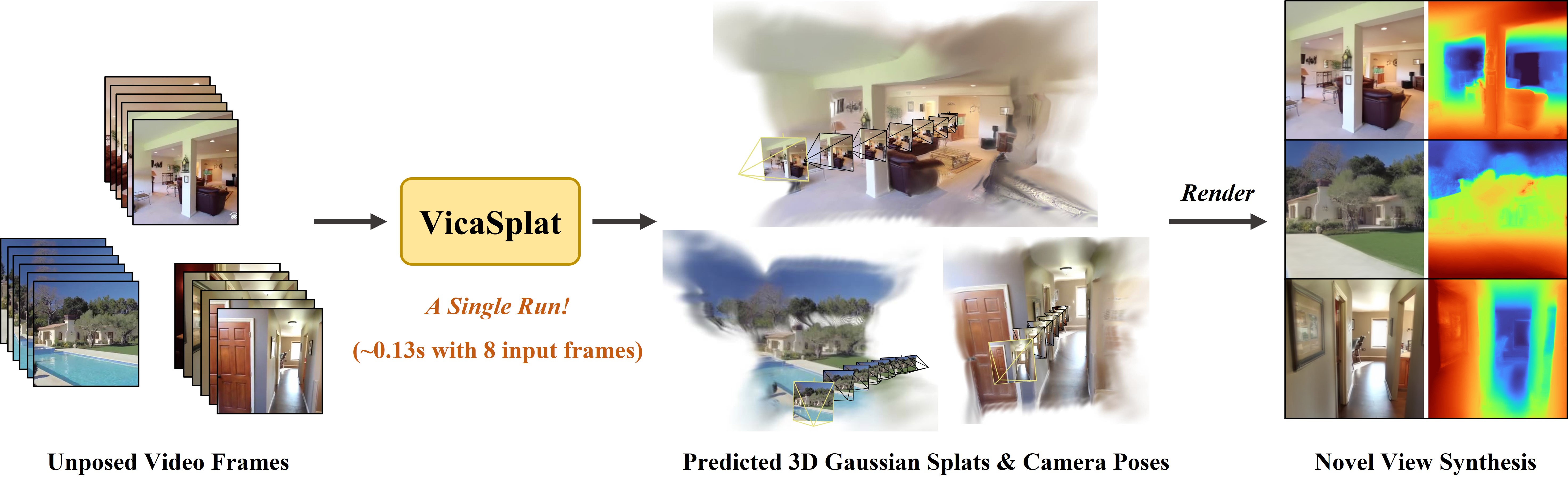}
           \vspace{-1em}
           \captionof{figure}{Provided with a sequence of unposed video frames, our method can predict high-quality 3D Gaussian Splatting and accurate camera poses in a single run (\eg 0.13 s for 8 input images).}
           \label{fig_teaser}
        \end{center}
    }]
}

\newcommand{\figref}[1]{Fig.~\ref{#1}}

\makeatletter
\DeclareRobustCommand\onedot{\futurelet\@let@token\@onedot}
\def\@onedot{\ifx\@let@token.\else.\null\fi\xspace}
\def\eg{e.g\onedot} 
\def\ie{i.e\onedot}

\makeatother

\newcommand{\PAR}[1]{\vspace{0.1cm}\noindent{\bf #1} }

\begin{document}


\maketitle
\let\thefootnote\relax\footnotetext{$^{\dag}$ Corresponding author.}

\begin{abstract}
We present \NAME, a novel framework for joint 3D Gaussians reconstruction and camera pose estimation from a sequence of unposed video frames, which is a critical yet underexplored task in real-world 3D applications.
The core of our method lies in a novel transformer-based network architecture. In particular, our model starts with an image encoder that maps each image to a list of visual tokens. All visual tokens are concatenated with additional inserted learnable camera tokens. The obtained tokens then fully communicate with each other within a tailored transformer decoder.
The camera tokens causally aggregate features from visual tokens of different views, and further modulate them frame-wisely to inject view-dependent features. 3D Gaussian splats and camera pose parameters can then be estimated via different prediction heads. Experiments show that {\name} surpasses baseline methods for multi-view inputs, and achieves comparable performance to prior two-view approaches. Remarkably, {\name} also demonstrates exceptional cross-dataset generalization capability on the ScanNet benchmark, achieving superior performance without any fine-tuning.
Codes and model weights are available at \url{https://lizhiqi49.github.io/VicaSplat}.
\end{abstract}

\section{Introduction}
\label{sec:intro}
Given only a sequence of video frames, human beings can easily infer the 3D spatial structure of the scene captured by the video, without the requirements of precise camera parameters. However, rapid 3D scene reconstruction and novel view synthesis (NVS) from unposed video frames still remains a difficult task in the field of computer vision.

Differentiable radiance fields ~\cite{mildenhall2021nerf,muller2022instant,kerbl20233dgaussian} have advanced novel view synthesis (NVS) using photometric loss from multi-view images, but require precise camera parameters from SfM ~\cite{schonberger2016colmap,wang2024vggsfm} and lengthy per-scene optimization. While some methods combine pose estimation with scene reconstruction ~\cite{lin2021barf,chen2023dbarf,hong2024coponerf}, they still need scene-specific optimization, limiting their feasibility for time-constrained applications.
Recent learning-based methods enable faster NVS without per-scene optimization, though most still need calibrated cameras~\cite{yu2021pixelnerf,chen2021mvsnerf,charatan2024pixelsplat,chen2024mvsplat}. Building on pre-trained point cloud models~\cite{wang2024dust3r,leroy2025mast3r}, which can reconstruct 3D scenes from unposed image pairs, some approaches now adapt this for Gaussian-based NVS without requiring camera poses~\cite{ye2024noposplat,smart2024splatt3r}. However, these methods still face efficiency challenges when processing multiple views, requiring extensive pairwise computations and global alignment.

\begin{figure}[tbp]
    \centering
    \includegraphics[width=\linewidth]{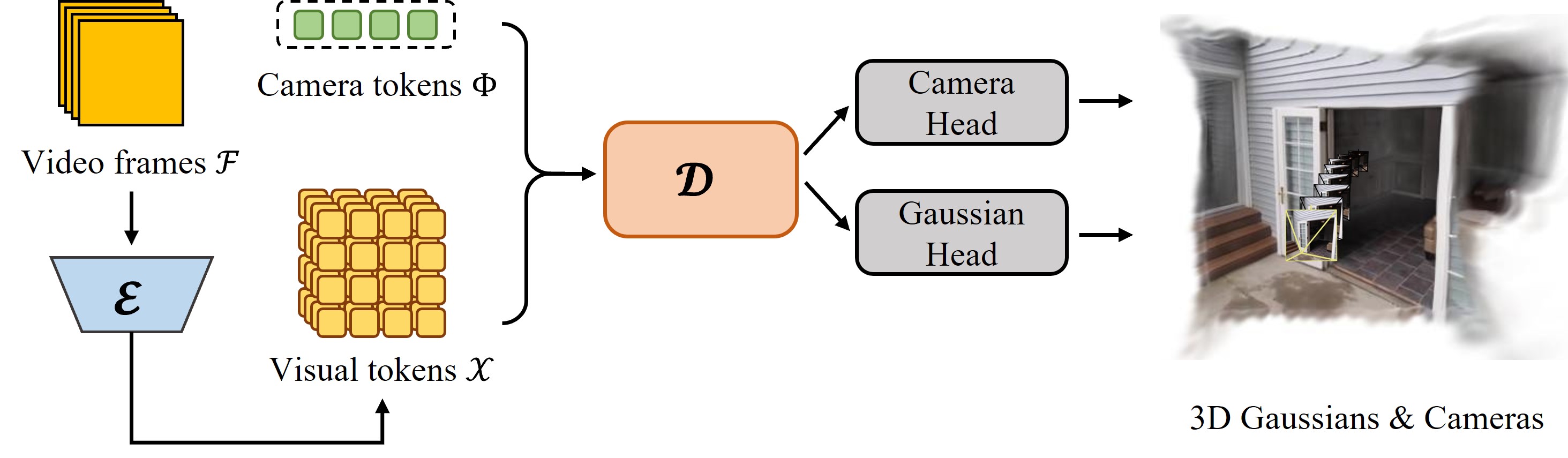}
    \caption{{\bf{Overview of the proposed method.}} The model employs a transformer encoder to convert video frames into visual tokens, while a custom transformer decoder with learnable camera tokens processes these representations. Two dedicated prediction heads then predict camera poses and 3D Gaussians respectively.}
    \label{fig_overview}
    \vspace{-1em}
\end{figure}

In order to overcome the aforementioned issues, we introduce \NAME, which predicts 3D Gaussian splats and camera poses from unposed video frames in one forward pass. As shown in \figref{fig_overview}, the model uses a transformer encoder to convert video frames into visual tokens, followed by a specialized decoder with learnable camera tokens. These camera tokens interact with visual tokens through bidirectional attention, enabling view-dependent feature extraction. The model then generates camera parameters and per-pixel 3D Gaussian splats through separate prediction heads. We represent camera poses using unit dual-quaternions for compactness and develop a new loss function for improved pose alignment.
Rather than training on extensive 3D datasets, we distill spatial knowledge from pretrained point cloud reconstruction models~\cite{wang2024dust3r, leroy2025mast3r} to save both the computation and storage budget. Our progressive training approach first focuses on scene geometry estimation, followed by joint optimization of novel view synthesis and camera pose regression. Experimental evaluations demonstrate that {\NAME} outperforms existing multi-view methods while matching the quality of two-view approaches, in terms of novel view synthesis. The model also exhibits robust generalization performance, achieving superior results on ScanNet without any fine-tuning.
Our main contributions can be summarized as follows: 
\begin{itemize}
\itemsep0em
\item We present an efficient feed-forward network for joint 3D Gaussian Splatting reconstruction and camera pose estimation from a sequence of video frames, without requiring the ground-truth camera poses;
\item We propose a novel transformer-based decoder network tailored to the input tokens, which is vital to the success of the algorithm;
\item The experimental results demonstrate our method surpasses prior methods for multi-view inputs, while achieving comparable performance to prior two-view approaches, in terms of novel view synthesis. The model also exhibits strong generalization performance on dataset without any fine-tuning.
\end{itemize}

\section{Related Work}
\label{sec:related}

\paragraph{Generalizable Novel View Synthesis.}
While Neural Radiance Fields (NeRF)~\cite{mildenhall2021nerf, sitzmann2019scene} and 3D Gaussian Splatting (3DGS)~\cite{kerbl20233dgaussian} have revolutionized novel view synthesis, they require numerous images with precise camera poses and lengthy per-scene optimization~\cite{chen2022tensorf, fridovich2023k, muller2022instant}. Recent methods attempt to jointly optimize camera poses and scene reconstruction~\cite{lin2021barf, wang2021nerf, bian2023nope, truong2023sparf, chen2023dbarf, hong2024coponerf,fan2024instantsplat, fu2024colmap, keetha2024splatam}, but they still need initial pose estimates. This interdependence can lead to error propagation between pose estimation and scene reconstruction, compromising overall performance.

To mitigate limitations brought by time-consuming per-scene optimization methods, recent learning-based approaches have extended NeRF and 3DGS for generalizable novel view synthesis from sparse images, demonstrating effective generalization to unseen scenes~\cite{yu2021pixelnerf, wang2021ibrnet, xu2023dmv3d, chen2024mvsplat, chen2021mvsnerf, chen2024mvsplat, zhang2024world}. By scaling up datasets and models, these methods achieve high-quality results without the per-scene optimization traditionally required by NeRF or 3DGS. These approaches incorporate specialized architectures with geometric priors to enhance reconstruction quality. Notable examples include MVSNeRF~\cite{chen2021mvsnerf}, MuRF~\cite{xu2024murf} and MVSplat~\cite{chen2024mvsplat}, which utilize cascade cost volume matching to aggregate multi-view information, while PixelSplat~\cite{charatan2024pixelsplat} leverages differentiable rendering and epipolar geometry for improved depth estimation. However, despite these advances in generalizable 3D reconstruction, these methods still require calibrated camera poses and sufficient view overlap between input images—conditions rarely met in practical scenarios. Camera poses typically need to be estimated using structure-from-motion (SfM) techniques~\cite{schonberger2016colmap,wang2024vggsfm}, which adds complexity to the inference pipeline. In contrast to these approaches where calibrated camera poses are essential throughout both training and inference, our method enables direct 3D reconstruction from sequences of uncalibrated images.

\paragraph{Feed-forward Multi-view Stereo Models.}
Traditional multi-view stereo (MVS) reconstruction typically follows a pipeline of depth map estimation and fusion before surface reconstruction~\cite{furukawa2015multi, Galliani_2015_ICCV}. DUSt3R~\cite{wang2024dust3r} and its successor MASt3R~\cite{leroy2025mast3r} revolutionize this approach by directly predicting pixel-aligned point maps from image pairs without requiring camera pose parameters, which are traditionally essential for maintaining photometric consistency. This innovation extends to dynamic scenes through MonST3R~\cite{zhang2024monst3r}, while NoPoSplat~\cite{ye2024noposplat} and Splatt3R~\cite{smart2024splatt3r} adapt this framework to predict 3D Gaussian primitives from image pairs, maintaining the pose-free and feed-forward characteristics of DUSt3R. However, these generalizable 3D Gaussian Splatting methods, built upon DUSt3R's architecture, face limitations in multi-view scenarios. While they can process multiple views through extensive pairwise inference followed by global feature alignment, this inefficient post-processing approach is time-consuming and often fails to fully integrate information across the input sequence, leading to inconsistencies between pair-wise predictions.

Our method addresses these limitations by providing a pose-free, feed-forward solution that eliminates the need for post-processing. Through an end-to-end network architecture, our approach processes unposed video sequences and simultaneously predicts frame-wise 3D Gaussian primitives and camera poses in a single forward pass. While Flare~\cite{zhang2025flare}, a concurrent work, also tackles pose-free multi-view Gaussian splatting prediction through multiple cascaded models, our end-to-end approach represents a fundamentally different solution.

\section{Method}
\label{sec:method}

\begin{figure*}[tbp]
    \centering
    \includegraphics[width=0.9\linewidth]{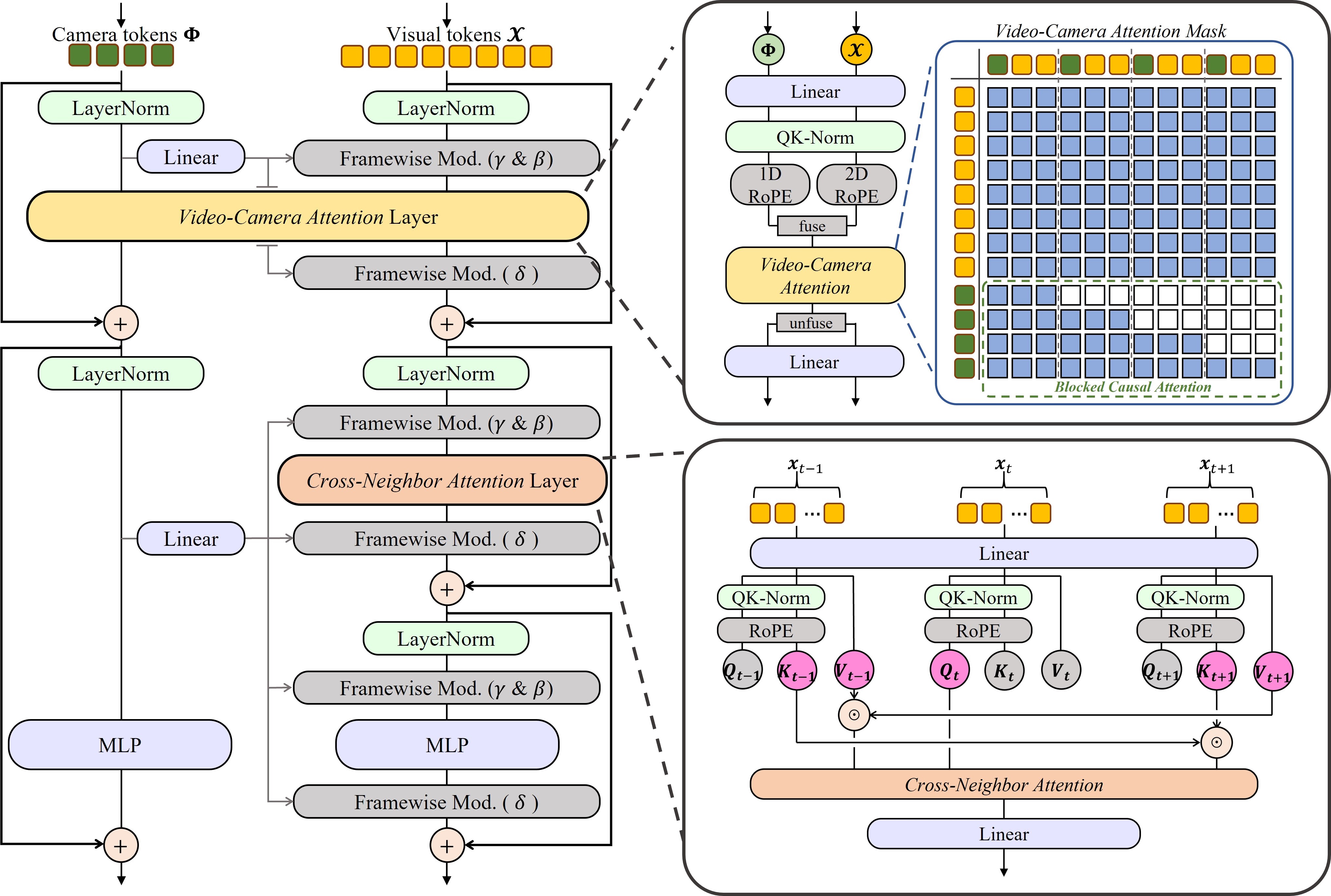}
    \vspace{-0.5em}
    \caption{{\bf{One block of {\name} decoder.}} }
    \label{fig_block}
    \vspace{-1em}
\end{figure*}

\subsection{Problem Formulation}
\label{subsec:problem_formulation}
Given a sequence of unposed RGB video frames $\mathcal{F}=\{\bm{f}_1, ...,\bm{f}_T\}$ of shape $T\times3\times H\times W$, where $\bm{f}_t\in\mathbb{R}^{3\times H\times W}$ is the $t$-th frame, we want our model to reconstruct the 3D scene by predicting pixel-aligned 3D Gaussians $\mathcal{G}=\{\cup(\bm{\mu}_i^t,\alpha_i^t,\bm{q}_i^t,\bm{s}_i^t,\bm{c}_i^t)\}_{i=1,...,H\times W}^{t=1,...,T}$ in the canonical 3D space (the camera coordinate space of the first frame) following previous works~\cite{wang2024dust3r, ye2024noposplat}. $\bm{\mu}\in\mathbb{R}^3$ is the Gaussian center position, $\alpha\in\mathbb{R}$ is the opacity, $\bm{q}\in\mathbb{R}^4$ is the rotation parameterized as quaternion, $\bm{s}\in\mathbb{R}^3$ is the scale, and $\bm{c}\in\mathbb{R}^k$ is spherical harmonics (SH) with $k$ degrees of freedom. Moreover, contrast to prior works where post-optimization process is required for camera pose estimation, our model is designed to predict the frame-wise camera poses $\mathcal{P}$ jointly with 3D Gaussians $\mathcal{G}$.

\subsection{Network Architecture}
\label{subsec:architecture}

The core of our approach lies in a transformer-based model~\cite{vaswani2017attention, dosovitskiy2020vit} consisting of an encoder $\mathcal{E}$ and a decoder $\mathcal{D}$. In particular, the encoder firstly maps each frame $\bm{f}_t$ into a set of visual tokens $\bm{x}_t\in\mathbb{R}^{\frac{H}{s}\times\frac{W}{s}\times C}$ where $s$ is the patch size and $C$ is the dimension of each token. The visual tokens of all frames, $\mathcal{X}=\{\bm{x}_1,\bm{x}_2,...,\bm{x}_T\}$, are then flattened and concatenated into a single token sequence with length $L=T\times\frac{H}{s}\times\frac{W}{s}$ and fed into the decoder $\mathcal{D}$. The architecture of $\mathcal{D}$ is tailored to accommodate the input of multiple video frames so that consistent scene-level 3D Gaussians, $\mathcal{G}$, and accurate camera extrinsic parameters $\mathcal{P}$ can be predicted from it.
In this section, we introduce the key components in the architecture of our decoder.



\paragraph{Learnable Camera Tokens.} In order to predict per-frame extrinsic parameters of the camera, we insert a set of learnable query tokens into the visual token sequence. Specifically, a learnable camera token $\bm{\phi}\in\mathbb{R}^C$ is expanded to shape $T\times C$ with 1D positional embedding applied to make them ordered.
The resulting $T$ camera tokens denoted as $\Phi=\{\bm{\phi}_1,...,\bm{\phi}_T\}$ are fed into the decoder $\mathcal{D}$ along with visual tokens $\mathcal{X}$. In the forward pass, the camera tokens fuse features from visual tokens of different frames, and further modulate the visual features.

\paragraph{Video-Camera Attention.} Within each block of our transformer decoder, the visual tokens and camera tokens firstly go through an attention layers where they fully communicate with each other. As shown in upper right of Fig. \ref{fig_block}, $\mathcal{X}$ and $\Phi$ are combined into a mixed sequence by prepending each camera token $\bm{\phi}_t$ to the corresponding visual tokens $\bm{x}_t$. The mixed sequence then conduct self-attention calculation, wherein the visual tokens share global scope, while a \emph{blocked causal attention} mask is applied to the camera tokens, so that each camera token cannot see neither the visual nor camera tokens after it. 1D and 2D rotary position embedding (RoPE)~\cite{su2024rope} are applied to camera tokens and visual tokens respectively.

\paragraph{Cross-Neighbor Attention.} After aggregating features from all visual and camera tokens, to enhance the view-consistency, we make the visual tokens further conduct cross-attention with their neighbor frames, namely \emph{cross-neighbor attention}. Firstly, the features of all visual tokens are mapped to their query $Q$, key $K$ and value $V$. Then for the $t$-th frame, the cross-neighbor attention (CNA) is calculated as:
\begin{equation}
\begin{aligned}
    \mathrm{CNA}_t=\mathtt{softmax}&(\frac{Q_t\tilde{K}_t^\mathrm{T}}{\sqrt{d_K}})\tilde{V}_t, 
    \\
    \tilde{K}_t=\mathtt{concat}(K_{t-1},K_{t+1}), &\tilde{V}_i=\mathtt{concat}(V_{t-1},V_{t+1}).
\end{aligned}
\end{equation}
Note that for the first frame and the last frame, they only do cross-attention with their next or previous one frame respectively. The lower right of Fig. \ref{fig_block} illustrate the process.

\paragraph{Framewise Modulation.} Once the camera tokens went through a video-camera attention layer, they would have been view-dependent due to their causally aggregating features from other tokens. Inspired by the way on how DiTs (Diffusion Transformers)~\cite{peebles2023dit} handles timestep embedding conditions, we propose to modulate the visual tokens with those differentiated camera tokens as control signals. 
In particular, before a specific layer denoted as $f(\cdot)$ in one of our attention block, we add a linear layer that regresses dimension-wise scale, shift and gate factor, $\gamma$, $\beta$ and $\delta$, for each frame of visual tokens, with their prepended camera token as input. The scale $\gamma$ and shift $\beta$ are applied before $f(\cdot)$, and the gate $\delta$ is utilized to control the strength of residual connection~\cite{he2016resnet} after $f(\cdot)$. With the $t$-th frame as example, the modulation procedure is formally:
\begin{align}
    \gamma_t,\beta_t,\delta_t&=\mathtt{Linear}(\bm{\phi}_t), \\
    \bm{x}_t=\bm{x}_t+(1+\delta_t&)(f(\bm{x}_t*(1+\gamma_t)+\beta_t)).
\end{align}
The linear layer is zero-initialized, so that the modulation operation does not affect the forward pass during the initial phase.
This process is illustrated with gray arrows in Fig. \ref{fig_block}. With our proposed frame-wise modulation, complex view-dependent features are injected into the visual tokens for better view-consistency learning.

\paragraph{Prediction Heads.} After the final transformer block, we use several prediction heads to obtain our desired targets. Specifically, we utilize a simple DPT head~\cite{ranftl2021dpt} to predict Gaussian centers $\bm{\mu}$, and a DPT-GS head~\cite{ye2024noposplat} for the other Gaussian features. Moreover, a linear head is employed to predict camera extrinsic parameters $\mathcal{P}$ from the camera tokens, which have aggregated rich view-dependent features. Note that we define the camera coordinate frame of the first image as the canonical space, its camera pose is thus always identical. Hence we only need to predict camera pose for those frames after the first one. 

\subsection{Camera Pose Regression}
\label{subsec:camera_loss}
\paragraph{Dual-Quaternion Parameterization.} Instead of directly regressing quaternion and translation vector for camera pose, we parameterize camera pose as a more compact representation, unit dual-quaternion (DQ), which has gained considerable interest in the past decades and is favored by robot modeling and control. There are two benefits of modeling camera pose as unit DQ: 1) DQ inherently couples rotation and translation into a single algebra, avoiding conflicting predictions from separate parameterizations \cite{daniilidis1999hand}, and making the regression easier for neural network; 2) Compared to independent rotation/translation interpolation, unit DQ interpolation can lead to more natural camera trajectory \cite{kavan2008dqs, kavan2006dual}, which matches well with the continuous property of neural networks.

\PAR{Align Prediction to Ground Truth.} Leveraging unit DQ's algebra that the production between an unit DQ and its conjugate leads to identical DQ $\hat{\bm{p}}$, we propose a novel loss term to align the predicted camera extrinsic parameters with the ground truth besides a simple regression loss term. In particular, for the predicted camera poses of all $T$ frames, $\mathcal{P}=\{\bm{p}_1,...,\bm{p}_T\}$, the proposed alignment loss and camera regression loss are calculated as:
\begin{equation}
    \mathcal{L}_{align}=\sum_{t=1}^{T}\lVert \hat{\bm{p}}-\bar{\bm{p}}_t\bm{p}_t^* \rVert + \sum_{t=1}^{T}\lVert \hat{\bm{p}}-\bm{p}_t\bar{\bm{p}}_t^*\rVert, 
\end{equation}
\begin{equation}
    \mathcal{L}_{camera}=\mathrm{MSE}(\mathcal{P},\mathcal{\bar{P}})+\mathcal{L}_{align},
\end{equation}
where $\bar{\bm{p}}_t\in\mathcal{\bar{P}}$ is the ground truth unit DQ for the $t$-th frame, and $\bm{p}^*$ is the conjugate of $\bm{p}$.

\subsection{Multi-view Training}
%


\paragraph{Progressive Multi-View Training.} While directly train our model with $N$ input views can be time-consuming and hard to converge, we adopt a progressive training strategy: we firstly train our model with two input views, during which stage the model learns the basic multi-view geometry knowledge and can predict high-quality 3D Gaussian features, then we gradually add input views until it reaches $N$ views. Within each training stage, we warm up the training by gradually enlarge the interval between the sampled video frames. Every time we increase the number of input views, the initial frame intervals are carefully selected so that the maximum gap from the first frame to the last is encapsulated by that of last training stage. With such a strategy, our model is pushed towards the desired capacity efficiently. 

\paragraph{3D Prior Distillation.} To construct our two-view base model, we need to learn a good geometry prior before applying for 3D Gaussian splatting prediction. Instead of training on the mixture of large-scale 3D datasets, we refer to distilling 3D knowledge from pretrained two-view 3D model~\cite{wang2024dust3r,leroy2025mast3r}. Given each pair of unposed images, $\bm{f}_a$ and $\bm{f}_b$, we supervise our model (via adding a new prediction head) with the point maps $\bar{Z}$ and confidence maps $\bar{C}$ predicted by the teacher model. For a valid pixel $i$ in view $v\in\{a, b\}$, the corresponding point distillation loss is computed as $\mathcal{L}(v,i)=||Z_i^{v,a}-\bar{Z}_i^{v,a}||$. We further sum up the loss on all valid pixels weighted with the predicted confidence maps:
\begin{equation}
    \mathcal{L}_{distill}=\sum_{v\in\{a,b\}}\sum_i\bar{C}_i^{v,a} \cdot \mathcal{L}(v,i) + ||C_i^{v,a}-\bar{C}_i^{v,a}||.
\end{equation}
Compared to the point regression loss in \cite{wang2024dust3r}, we omit the normalization term to straightforwardly imitate behavior of the teacher model. Note that we also make our model to predict the confidence map in point distillation stage. It helps the model to understand the varying importance of different regions in given scenes. Once we have fully distilled the teacher model, we start our progressive multi-view training for novel view synthesis and camera pose estimation.

\paragraph{Total Loss.} The total loss to train our model is given below:
\begin{equation}
    \mathcal{L}=\begin{cases}
     \mathcal{L}_{distill}, &\text{during 3D distillation phase} \\
     \mathcal{L}_{img}+\lambda\mathcal{L}_{camera}, &\text{otherwise} \\
     \end{cases}
\end{equation}
where $\mathcal{L}_{img}$ is the normal photometric loss which is a linear combination of MSE and LPIPS~\cite{zhang2018lpips} loss with weights of 1 and 0.05 respectively, and $\lambda=0.1$ is the weight for camera pose regression loss defined in Sec. \ref{subsec:camera_loss}.
\section{Experiments}
\label{sec:exp}


\begin{figure*}\label{fig_results}
	\centering
	\includegraphics[width=0.95\linewidth]{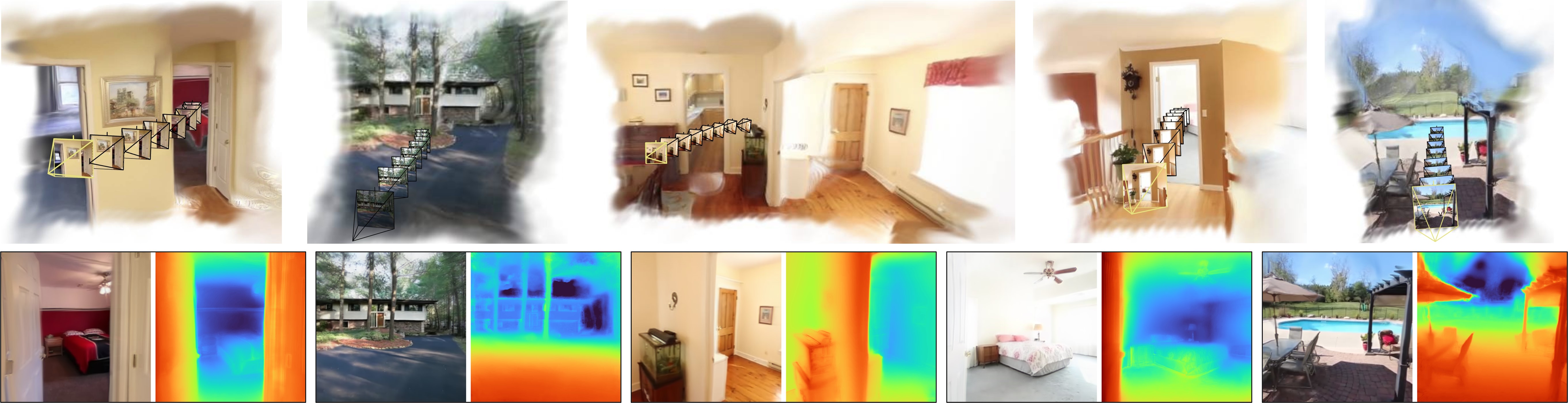}
	\vspace{-0.5em}
	\captionof{figure}{{\bf{Predicted 3D Gaussian splats and camera poses (top row) as well as rendered novel view RGBs and depths (bottom row).}} Our model can jointly reconstruct 3D Gaussians and recover camera extrinsic parameters through a single forward pass. High-fidely RGB images and depth maps can be rendered from novel views.}
        \vspace{-1em}
\end{figure*}

\paragraph{Implementation Details.} Our encoder is a 24-layers ViT model with width as 1024, which is the same as DUSt3R~\cite{wang2024dust3r} and MASt3R~\cite{leroy2025mast3r}, and we initialize its weights with MASt3R's encoder.
As for our decoder, we design it with 12 layers of our proposed Video-Camera Attention block, with hidden size of 768. The framewise modulation layers are zero-initialized to avoid harming the forward pass of the visual part at beginning. Our model is implemented with PyTorch~\cite{paszke2019pytorch} and trained with images at resolution of $256\times256$ pixels, with the maximum input length as $N=8$. All our models are trained with $8\times$A100 GPUs. Our 8-view model takes $\sim0.13$s for inference on a single A100 GPU with peak memory consumption as $\sim8$GB.

\begin{figure*}
    \centering
    \addtolength{\tabcolsep}{-6.5pt}
    \footnotesize{
        \setlength{\tabcolsep}{1pt} 
        \begin{tabular}{p{8.2pt}ccccc}
            & Scene1 & Scene2 & Scene3 & Scene4 & Scene5  \\
            
        \raisebox{36pt}{\rotatebox[origin=c]{90}{\textbf{PixelSplat}}}&
         \includegraphics[width=0.15\textwidth]{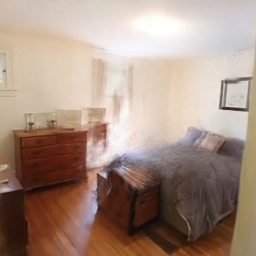} &
        \includegraphics[width=0.15\textwidth]{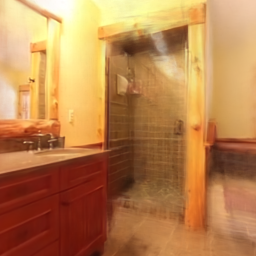} &
        \includegraphics[width=0.15\textwidth]{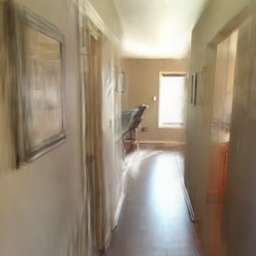} &
        \includegraphics[width=0.15\textwidth]{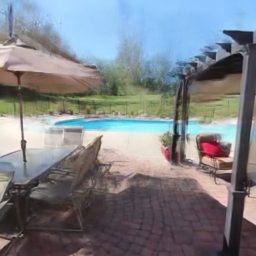} &
        \includegraphics[width=0.15\textwidth]{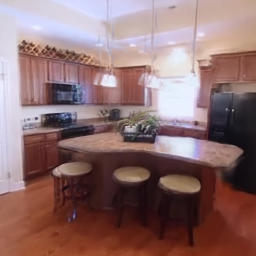} \\

        \raisebox{36pt}{\rotatebox[origin=c]{90}{\textbf{MVSplat}}}&
         \includegraphics[width=0.15\textwidth]{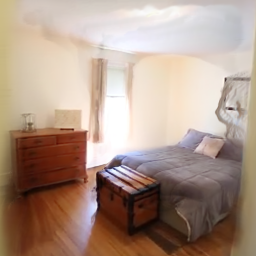} &
        \includegraphics[width=0.15\textwidth]{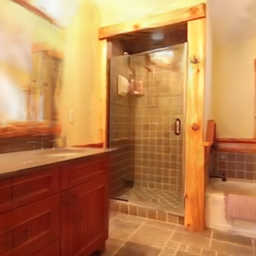} &
        \includegraphics[width=0.15\textwidth]{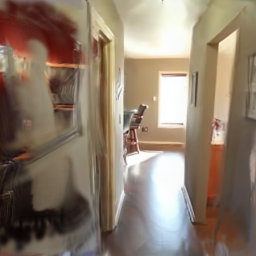} &
        \includegraphics[width=0.15\textwidth]{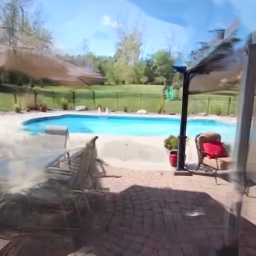} &
        \includegraphics[width=0.15\textwidth]{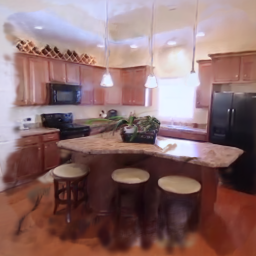}  \\

        \raisebox{36pt}{\rotatebox[origin=c]{90}{\textbf{FreeSplat}}}&
         \includegraphics[width=0.15\textwidth]{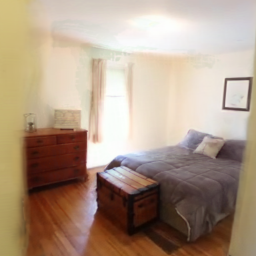} &
        \includegraphics[width=0.15\textwidth]{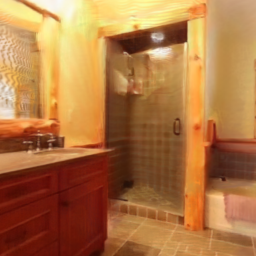} &
        \includegraphics[width=0.15\textwidth]{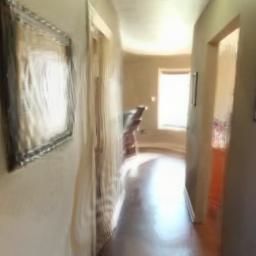} &
        \includegraphics[width=0.15\textwidth]{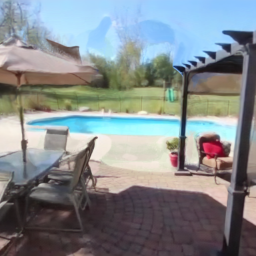} &
        \includegraphics[width=0.15\textwidth]{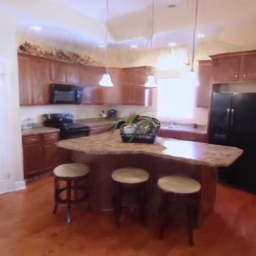} \\

        \raisebox{36pt}{\rotatebox[origin=c]{90}{\textbf{NoPoSplat}}}&
         \includegraphics[width=0.15\textwidth]{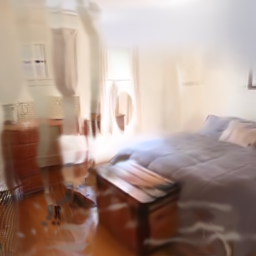} &
        \includegraphics[width=0.15\textwidth]{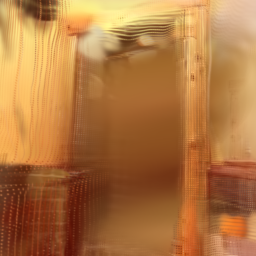} &
        \includegraphics[width=0.15\textwidth]{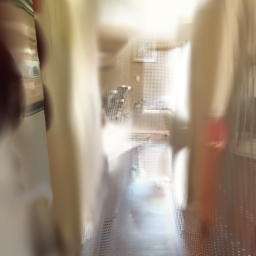} &
        \includegraphics[width=0.15\textwidth]{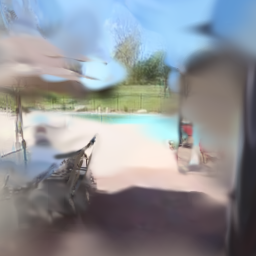} &
        \includegraphics[width=0.15\textwidth]{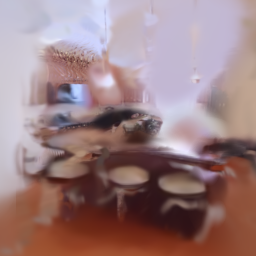} 
         \\
         
        \raisebox{36pt}{\rotatebox[origin=c]{90}{\textbf{Ours}}}&
         \includegraphics[width=0.15\textwidth]{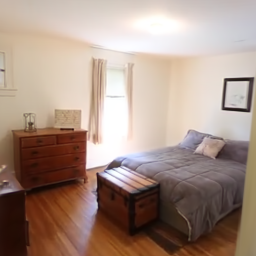} &
        \includegraphics[width=0.15\textwidth]{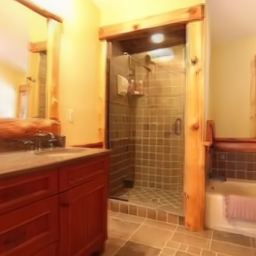} &
        \includegraphics[width=0.15\textwidth]{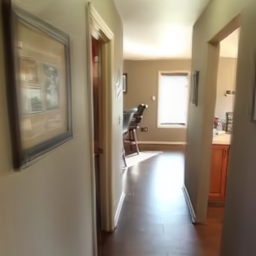} &
        \includegraphics[width=0.15\textwidth]{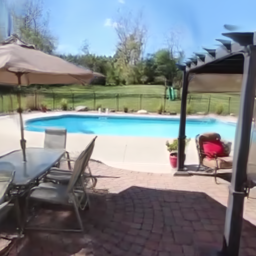} &
        \includegraphics[width=0.15\textwidth]{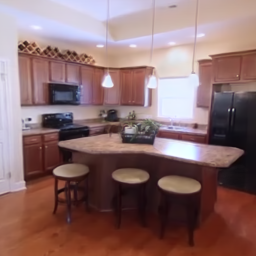} 
         \\
        
        \raisebox{36pt}{\rotatebox[origin=c]{90}{\textbf{Ground Truth}}}&
         \includegraphics[width=0.15\textwidth]{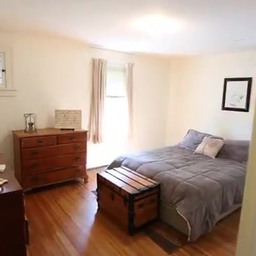} &
        \includegraphics[width=0.15\textwidth]{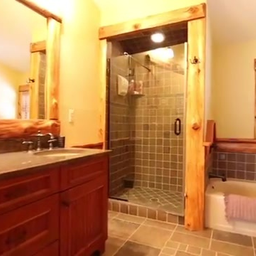} &
        \includegraphics[width=0.15\textwidth]{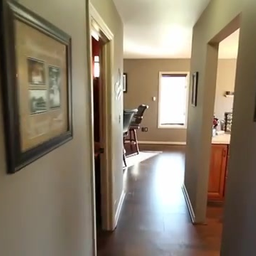} &
        \includegraphics[width=0.15\textwidth]{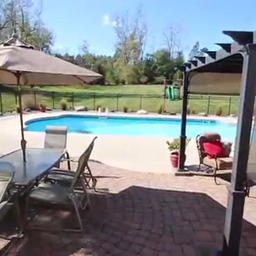} &
        \includegraphics[width=0.15\textwidth]{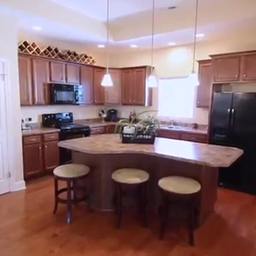} 
         \\

        \end{tabular}
    }
    \vspace{-1.5pt}
    \caption{{\bf{Qualitative comparison of novel view synthesis on RealEstate10k test set with 8 input images.}} Our model produces the best rendering details and geometry accuracy compared to all baseline methods.}
    \label{fig_comp_8view}
    \vspace{-1em}
\end{figure*}

\begin{table}[!t]
    \centering
    \begin{adjustbox}{max width=\linewidth}
    \setlength{\tabcolsep}{0.08cm}
    \renewcommand{\arraystretch}{1.0}
    \begin{tabular}{llccc | ccc}
        \toprule
        & & \multicolumn{3}{c}{RE10k} & \multicolumn{3}{c}{ACID} \\ 
        \cmidrule(lr){3-5} \cmidrule(lr){6-8}
        & Method & PSNR$\uparrow$ & SSIM$\uparrow$ & LPIPS$\downarrow$ & PSNR$\uparrow$ & SSIM$\uparrow$ & LPIPS$\downarrow$ \\
        \midrule
        \multirow{4}{*}{\shortstack[l]{\emph{Pose-} \\ \emph{Required}}} 
        & pixelSplat & 23.361 & 0.803 & 0.186 & \second{25.684} & \second{0.778} & \second{0.194} \\
        & MVSplat & 23.430 & 0.805 & 0.179 & 25.335 & 0.772 & 0.195 \\
        & FreeSplat & 16.403 & 0.541 & 0.452 & 17.903 & 0.548 & 0.438 \\
        \midrule
        \multirow{5}{*}{\shortstack[l]{\emph{Pose-} \\ \emph{Free}}} 
        & NoPoSplat & \second{25.036} & \best{0.838} & \second{0.162} & \best{25.961} & \best{0.781} & \best{0.189} \\
        & CoPoNeRF & 18.938 & 0.619 & 0.388 & 20.950 & 0.606 & 0.406 \\
        & Ours(w/o intrin) & 24.053 & 0.799 & 0.176 & 24.548 & 0.726 & 0.218 \\
        & \best{Ours} & \best{25.038} & \second{0.834} & \best{0.161} & 25.439 & 0.757 & 0.201 \\
        \bottomrule
    \end{tabular}
    \end{adjustbox}
    \vspace{-0.5em}
    \caption{\textbf{Quantitative comparisons of novel view synthesis on the RealEstate10k~\citep{re10k} and ACID~\citep{acid} dataset under 2-views setup}. 
    Our method achieves comparable performance prior state-of-the-art method on both datasets, with two-view images. }
    \vspace{-1.5em}
\label{tab:re10k_acid_2view}
\end{table}

\begin{table*}[!t]
    \centering
    \footnotesize
    \begin{adjustbox}{max width=\textwidth}
    \setlength{\tabcolsep}{0.08cm}
    \renewcommand{\arraystretch}{1.0}
    \begin{tabular}{llccc ccc | ccc ccc}
        \toprule
        & & \multicolumn{3}{c}{4 views (RE10k)} & \multicolumn{3}{c}{8 views (RE10k)} & \multicolumn{3}{c}{4 views (ScanNet)} & \multicolumn{3}{c}{8 views (ScanNet)} \\
        \cmidrule(lr){3-5} \cmidrule(lr){6-8} \cmidrule(lr){9-11} \cmidrule(lr){12-14}
        & Method & PSNR$\uparrow$ & SSIM$\uparrow$ & LPIPS$\downarrow$ & PSNR$\uparrow$ & SSIM$\uparrow$ & LPIPS$\downarrow$ & PSNR$\uparrow$ & SSIM$\uparrow$ & LPIPS$\downarrow$ & PSNR$\uparrow$ & SSIM$\uparrow$ & LPIPS$\downarrow$ \\
        \midrule
        \multirow{4}{*}[0.7em]{\centering \shortstack[l]{\emph{Pose-} \\ \emph{Required}}} 
        & pixelSplat & 20.459 & 0.729 & 0.267 & 19.734 & 0.694 & \second{0.290} & 21.185 & 0.712 & 0.351 & 18.582 & 0.637 & 0.440 \\
        & MVSplat & \second{20.882} & \second{0.761} & \second{0.233} & 19.726 & 0.743 & 0.262 & 14.946 & 0.472 & 0.544 & 13.061 & 0.407 & 0.608 \\
        & FreeSplat & 20.630 & 0.747 & 0.313 & \second{21.259} & \second{0.767} & 0.309 & \best{28.304} & \second{0.849} & \second{0.205} & \best{27.040} & \best{0.832} & \best{0.224}  \\
        \midrule
        \multirow{5}{*}[1em]{\centering \shortstack[l]{\emph{Pose-} \\ \emph{Free}}} 
        & NoPoSplat & 16.299 & 0.552 & 0.397 & 14.372 & 0.469 & 0.520 & 16.940 & 0.560 & 0.425 & 12.939 & 0.397 & 0.588 \\
        & CoPoNeRF & 18.299 & 0.655 & 0.559 & 18.984 & 0.692 & 0.553 & 20.247 & 0.775 & 0.535 & 19.821 & 0.772 & 0.542 \\
        & \best{Ours} & \best{24.537} & \best{0.814} & \best{0.162} & \best{24.502} & \best{0.806} & \best{0.164} & \second{26.673} & \best{0.856} & \best{0.188} & \second{23.656} & \second{0.777} & \second{0.262} \\
        \bottomrule
    \end{tabular}
    \end{adjustbox}
    \vspace{-0.5em}
    \caption{\textbf{Novel view synthesis comparison with 4-view and 8-view input on the RE10k~\citep{re10k} and ScanNet~\citep{scannet} datasets}. 
    Our method outperforms previous baseline methods on RE10k. It also exhibits strong generalization capability on ScanNet even not being specifically trained on this dataset. Note that FreeSplat has been trained with ScanNet.}
    \vspace{-1em}
\label{tab:re10k_scannet_mv}
\end{table*}

\subsection{Novel-View Synthesis}
\paragraph{Dataset.} We evaluate novel view synthesis (NVS) with different numbers of input views. Under two-view setup, we follow the setting in \cite{charatan2024pixelsplat, chen2024mvsplat, ye2024noposplat} and evaluate the rendering quality on RealEstate10k (RE10k)~\cite{re10k} and ACID~\cite{acid} respectively. RE10k primarily contains indoor walkthrough videos collected from YouTube, while ACID consists of aerial landscape videos. We use the same test split as NoPoSplat~\cite{ye2024noposplat} on these two benchmarks. Then we extend our model to 4 and 8 input views on RE10k, with the test cases as totally 500 pieces of videos randomly sampled from 2-view test split, in avoid of too long evaluation time. Moreover, after training on RE10k, we evaluate our model's cross-dataset generalization capability on ScanNet~\cite{scannet} with the same number of input views. We randomly sample 100 scenes from ScanNet for testing.


\paragraph{Baselines and Metrics.}
We evaluate NVS with common photo metrics: PSNR, SSIM and LPIPS~\cite{zhang2018lpips}. Existing baselines including pose-required method, PixelSplat~\cite{charatan2024pixelsplat}, MVSplat~\cite{chen2024mvsplat} and FreeSplat~\cite{wang2024freesplat}, as well as pose-free methods, such as CoPoNeRF~\cite{hong2024coponerf} and NoPoSplat~\cite{ye2024noposplat},  are primarily designed for two-view inputs except FreeSplat. On two-view experiments, we benchmark all methods using their default configurations. When extending to 4-view and 8-view experiments, we make the epipolar attention only conducted on neighboring frames for PixelSplat, and apply DUSt3R-like global alignment~\cite{wang2024dust3r} for NoPoSplat.
Following NoPoSplat, we also utilize camera intrinsic conditioning and evaluate with pose alignment for novel view synthesis. 

\paragraph{Comparison on Two Views.} As reported in Tab. \ref{tab:re10k_acid_2view}, on RE10k, our model achieves comparable performance with NoPoSplat, which is the SOTA pose-free method. VicaSplat also significantly outperforms another pose-free method, CoPoNeRF. Notably, our method surpasses the pose-required SOTA methods, PixelSplat and MVSplat, with about 1.6dB on PSNR. Moreover, VicaSplat also reveals competitive performance on ACID, with only a gap of $\sim$0.5dB compared to the highest score on PSNR. We also report the scores of our model without intrinsic embedding. It has a drop of nearly 1dB on both dataset, but is more flexible for real-world application. In summary, VicaSplat can achieve competitive performance to state-of-the-art methods on both datasets in terms of novel view synthesis. 

\paragraph{Comparison on Multiple Views.} As demonstrated in Tab. \ref{tab:re10k_scannet_mv} and Fig. \ref{fig_comp_8view}, our model significantly outperforms all compared baselines on RE10k with both 4 or 8 views as input. When being evaluated on ScanNet, {\name} still achieves high-quality novel view rendering even it is not trained on it, with PSNR of 26.67 and 23.66 on 4-view and 8-view input respectively. It surpass all baseline methods except FreeSplat, which is trained on ScanNet. The results show the superior performance of our method on multi-view images, as well as its notable generalization performance.

\begin{table*}[t]
    \centering
    \footnotesize
    \begin{adjustbox}{max width=\textwidth}
    \setlength{\tabcolsep}{0.08cm}
    \renewcommand{\arraystretch}{1.0}
    \begin{tabular}{ll ccc ccc | ccc ccc}
        \toprule
        & & \multicolumn{3}{c}{4 views (RE10k)} & \multicolumn{3}{c}{8 views (RE10k)} & \multicolumn{3}{c}{4 views (ScanNet)} & \multicolumn{3}{c}{8 views (Scannet)} \\
        \cmidrule(lr){3-5} \cmidrule(lr){6-8} \cmidrule(lr){9-11} \cmidrule(lr){12-14}
        & Method & ATE$\downarrow$ & RPE-trans$\downarrow$ & RPE-rot$\downarrow$ & ATE$\downarrow$ & RPE-trans$\downarrow$ & RPE-rot$\downarrow$ & ATE$\downarrow$ & RPE-trans$\downarrow$ & RPE-rot$\uparrow$ & ATE$\downarrow$ & RPE-trans$\downarrow$ & RPE-rot$\uparrow$ \\
        \midrule
        & DUSt3R~\cite{wang2024dust3r} & 0.057 & 0.118 & 2.636 & 0.074 & 0.091 & 1.573 & 0.441 & 0.906 & 1.221 & 1.170 & 1.621 & 1.326 \\
        & MASt3R~\cite{leroy2025mast3r} & 0.064 & 0.129 & 2.753 & 0.090 & 0.117 & 1.642 & 0.365 & 0.858 & \second{1.241} & 0.900 & 1.332 & \second{1.241} \\
        & MonST3R~\cite{zhang2024monst3r} & 0.048 & 0.113 & 4.477 & 0.054 & 0.068 & 2.225 & 0.182 & 0.431 & 1.278 & 0.533 & 0.762 & 1.926 \\
        & VGGSFM~\cite{wang2024vggsfm} & 0.024 & 0.056 & 3.845 & \best{0.016} & \best{0.018} & 1.108 & 0.192 & 0.449 & 1.328 & 0.393 & 0.550 & 1.381 \\
        & Spann3R~\cite{wang2024spann3r} & 0.041 & 0.098 & 4.160 & 0.049 & 0.061 & 1.424 & 0.179 & 0.393 & \best{0.715} & 0.411 & 0.530 & \best{0.804} \\
        \midrule
        & \best{Ours} & \second{0.020} & \second{0.046} & \second{1.129} & 0.021 & 0.031 & \second{0.793} & \second{0.089} & \second{0.222} & 2.173 & \best{0.136} & \best{0.173} & 2.908 \\
        & \best{Ours$^*$} & \best{0.018} & \best{0.040} & \best{1.114} & \second{0.019} & \second{0.029} & 
        \best{0.689} & \best{0.085} & \best{0.200} & 1.760 & \second{0.151} & \second{0.210} & 3.692 \\
        \bottomrule
    \end{tabular}
    \end{adjustbox}
    \vspace{-0.5em}
    \caption{\textbf{Camera Pose Estimation on RE10k~\cite{re10k} and ScanNet~\cite{scannet}.} Our method achieves superior pose estimation performance under both 4-view and 8-view setup on RE10k, and can also predict accurate camera poses on ScanNet even it is not trained on this dataset.}
    \vspace{-1.5em}
\label{tab:pose_est}
\end{table*}

\subsection{Multi-View Camera Pose Estimation}
\paragraph{Dataset.} To evaluate pose estimation performance, we train our model on the RE10k dataset~\cite{re10k} and conduct testing on both RE10k and ScanNet~\cite{scannet}. For RE10k, we follow the same test split as NoPoSplat~\cite{ye2024noposplat} and randomly select 300 samples from this split for evaluation. For ScanNet, we utilize the data split employed in FreeSplat~\cite{wang2024freesplat}, comprising approximately 100 scenes. During testing, we randomly sample 4 or 8 views from each video, varying the sampling intervals to provide diverse input configurations for the model.

\paragraph{Baselines and Metrics.} We compare our method with recent 3R-series approaches, including DUSt3R~\cite{wang2024dust3r}, MASt3R~\cite{leroy2025mast3r}, MonST3R~\cite{zhang2024monst3r}, and Spann3R~\cite{wang2024spann3r}. Additionally, we include the state-of-the-art deep optimization-based method, \ie VGGSFM~\cite{wang2024vggsfm}, in our comparisons. For evaluation, we compute the Absolute Translation Error (ATE), Relative Translation Error (RPE-trans), and Relative Rotation Error (RPE-rot). To establish a consistent reference frame, we define the first camera coordinate system as the canonical space and transform all ground truth and estimated camera poses into this frame. Furthermore, we normalize the camera trajectory by a scaling factor so that the translation of the last camera pose has a unit norm. 

For our method, we report results of both directly predicted poses (\textbf{ours}) and that after further optimization-based alignment (\textbf{ours}$^*$) following NoPoSplat~\cite{ye2024noposplat}.

\paragraph{Comparison.}
As demonstrated in Table~\ref{tab:pose_est}, our method achieves superior performance compared to both optimization-based and online methods on RE10k~\cite{re10k} dataset. Notably, even on the ScanNet dataset~\cite{scannet}, where our model was not explicitly trained, our approach outperforms most baselines, including those trained on ScanNet (e.g., DUSt3R~\cite{wang2024dust3r}). Moreover, as the number of input views increases from 4 to 8, our method achieves comparable results on ScanNet and better performance on RE10k. This proves the effectiveness of information exchange between visual tokens and camera tokens in our model, which leads to better alignment in poses. 

\begin{figure}[tbp]
    \centering
    \includegraphics[width=\linewidth]{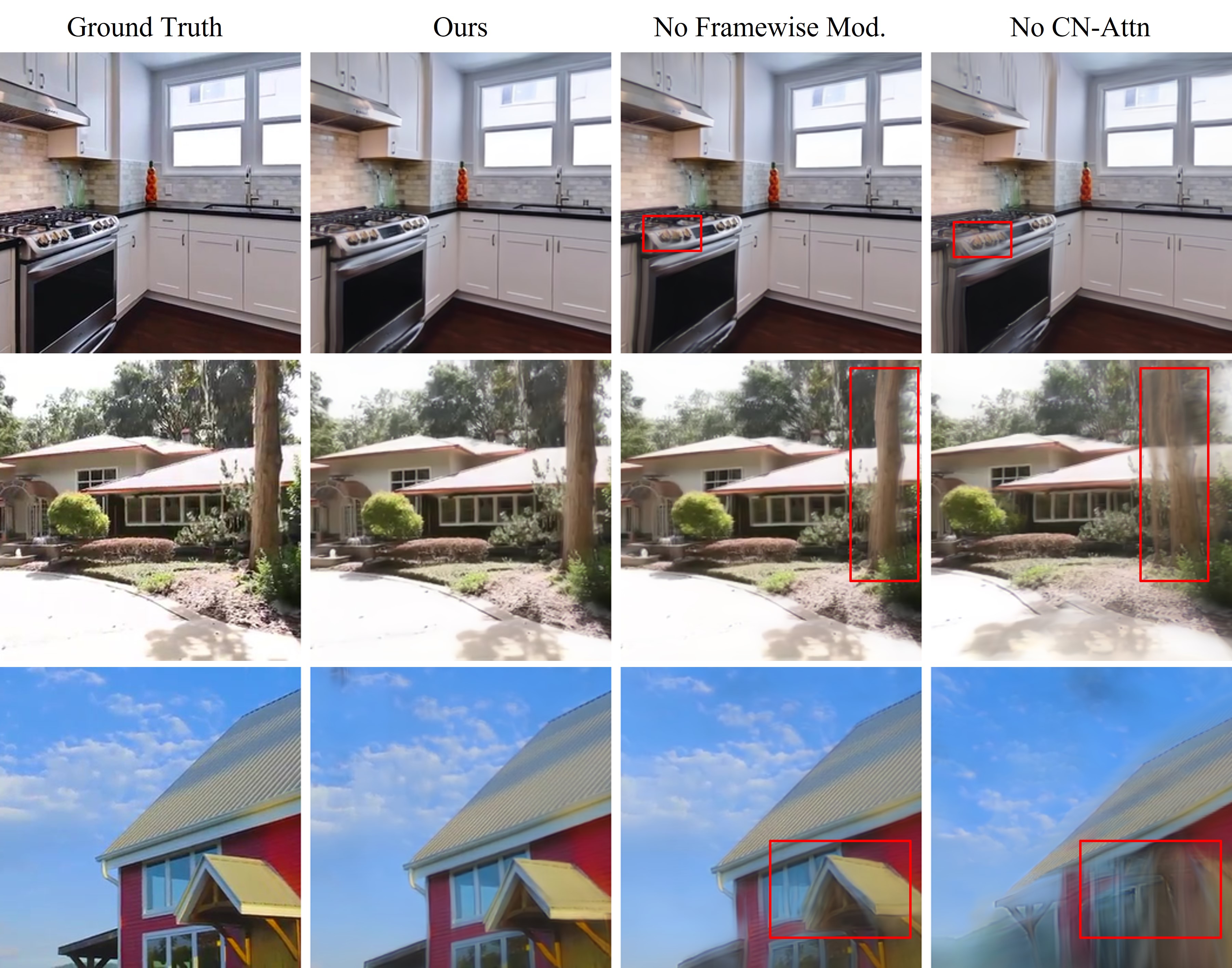}
    \vspace{-1em}
    \caption{{\bf{Ablations on novel view synthesis.}} Without our proposed framewise modulation and cross-neighbor attention layer, there is a obvious degeneration on the geometry prediction, and it presents severe blur in the rendered images.}
    \label{fig_abl_nvs}
    \vspace{-1.2em}
\end{figure}

\begin{table}[t]
\centering
\begin{adjustbox}{max width=\linewidth}
\setlength{\tabcolsep}{0.08cm}
\renewcommand{\arraystretch}{1.0}
\begin{tabular}{ll ccc ccc}
\toprule
& & \multicolumn{3}{c}{Novel-view Synthesis} & \multicolumn{3}{c}{Pose Estimation} \\
\cmidrule(lr){3-5} \cmidrule(lr){6-8}
Num & Design & PSNR & SSIM & LPIPS & ATE & RPE trans & RPE rot \\
\midrule
(a) & Ours & \best{22.516} & \second{0.738} & \best{0.197} & \second{0.020} & \second{0.046} & \best{1.129} \\
\midrule
(b) & No framewise mod. & 21.655 & 0.706 & 0.222 & 0.026 & 0.060 & 1.616  \\
(c) & No CN-attn & 19.908 & 0.644 & 0.287 & 0.042 & 0.107 & 2.691 \\
(d) & full attn. for camera & 22.505 & 0.738 & 0.201 & 0.021 & 0.048 & \second{1.240} \\
(e) & No DQ alignment loss & \second{22.514} & 0.738 & \best{0.197} & \best{0.019} & \best{0.045} & 1.264 \\
(f) & Quat.+Trans. for camera & 22.513 & \best{0.740} & \best{0.197} & \best{0.019} & \second{0.046} & 1.314 \\
\bottomrule
\end{tabular}
\end{adjustbox}
\vspace{-1em}
\caption{\textbf{Ablations.} Evaluation-time pose alignment is not conducted for clear comparison. Both framewise modulation and cross-neighbor attention are critical for the NVS. The blocked causal attention, DQ paramterization for camera pose,  and the novel DQ alignment loss all aid the model to learn better camera orientation. Camera pose estimation is essentially based on geometry learning, as evidenced by row (b) and row (c).
}
\vspace{-1em}
\label{tab:ablation}
\end{table}
\subsection{Ablation Study} We perform ablations on RE10k~\cite{re10k} with 4 input views to evaluate the effectiveness of our key components. We did not conduct evaluation-time pose alignment for NVS comparison. We report the quantitative results on both novel view synthesis and camera pose estimation in Table \ref{tab:ablation}.  




\paragraph{What is important for NVS?} As discussed in Sec. \ref{subsec:architecture}, we introduce frame-wise modulation to make the visual tokens view-dependent, and insert an additional cross-neighbor attention layer to enhance view-consistency. Both of them are critical for novel view synthesis. Without frame-wise modulation, a PSNR drop of nearly 0.9 dB is observed. When further removing the cross-neighbor attention layers, we witness a more severe performane decrease, which is $\approx1.7$ dB on PSNR. Qualitatively, as shown in Fig. \ref{fig_abl_nvs}, severe motion blur artifacts arise from inconsistencies in the predicted Gaussian centers across different views.

\paragraph{What is important for pose estimation?} As is shown in the Table \ref{tab:ablation}, what the most important to pose estimation is the learning of novel view synthesis, or spatial geometry. When the geometry is bad as shown in row (a) and row (b), the performance also decreases a lot on pose estimation metrics. In addition, the scores in row (d), row(e) and row (f) reveal that, blocked causal attention for camera tokens, camera pose DQ parameterization and DQ alignment loss help the model to learn better camera orientation, as evidenced by the increase on Relative Rotation Error.

\section{Conclusion}
\label{sec:conclusion}
We present {\name}, a fast feed-forward transformer-based network for joint 3D Gaussian splats prediction and camera pose estimation, from a sequence of unposed video frames. The proposed components, framewise modulation and cross-neighbor attention, effectively boost the geometry learning. By parameterizing camera poses as dual-quaternions, a novel camera alignment loss is further proposed to better align the predicted camera orientation to the ground truth. Thanks to the tailored model architecture, we push the model towards a superior performance with an efficient progressive training strategy.
The experiments demonstrate that our model achieves comparable capability against SOTA novel view synthesis models on two-views setting. When given multiple views, our model significantly outperforms baseline methods and reveals robust cross-dataset generalization capability in terms of both novel view synthesis and camera pose estimation.

{
    \small
    \bibliographystyle{ieeenat_fullname}
    \bibliography{reference}
}


\end{document}